\documentclass[a4paper,conference]{IEEEtran}
\ifCLASSINFOpdf
 \usepackage[pdftex]{graphicx}
 \graphicspath{{./images/}}
\else
\fi
\usepackage{amsmath}
\usepackage{amsfonts}
\usepackage{algorithmic}

\usepackage{array}

\ifCLASSOPTIONcompsoc
  \usepackage[caption=false,font=normalsize,labelfont=sf,textfont=sf]{subfig}
\else
  \usepackage[caption=false,font=footnotesize]{subfig}
\fi
\usepackage{url}

\usepackage{color}

\begin{document}
%
\title{Effects of sampling skewness of the importance-\\weighted risk estimator on model selection.}

\author{\IEEEauthorblockN{Wouter Kouw}
\IEEEauthorblockA{Netherlands eScience Center\\
Science Park 140, 1098 XG Amsterdam, The Netherlands\\
Email: w.kouw@esciencecenter.nl}
\and
\IEEEauthorblockN{Marco Loog}
\IEEEauthorblockA{Delft University of Technology\\
Mekelweg 4, 2628 CD, Delft, The Netherlands\\
University of Copenhagen\\
Universitetsparken 5, Copenhagen DK-2100, Denmark\\
Email: M.Loog@tudelft.nl}}


%


\maketitle

\begin{abstract}
Importance-weighting is a popular and well-researched technique for dealing with sample selection bias and covariate shift. It has desirable characteristics such as unbiasedness, consistency and low computational complexity. However, weighting can have a detrimental effect on an estimator as well. In this work, we empirically show that the sampling distribution of an importance-weighted estimator can be skewed. For sample selection bias settings, and for small sample sizes, the importance-weighted risk estimator produces overestimates for datasets in the body of the sampling distribution, i.e. the majority of cases, and large underestimates for data sets in the tail of the sampling distribution. These over- and underestimates of the risk lead to suboptimal regularization parameters when used for importance-weighted validation. 
\end{abstract}


%
\IEEEpeerreviewmaketitle

\section{Introduction}
Sampling with selection bias is often the only means to acquire data. \emph{Bias} in this context refers to the fact that certain observations occur more frequently than normal \cite{heckman1990varieties}. For instance, in social science experiments, data collected from university students will have different properties than data collected from the larger national or global population. This results in a statistical classification problem where the training and test data come from different distributions. Such problems are very challenging, because the information that is relevant to accurately classify training samples might not be relevant to classify test samples. This problem is more commonly known as \emph{sample selection bias} or \emph{covariate shift} \cite{cortes2008sample,quionero2009dataset,moreno2012unifying}. The setting from which the training data originates is often referred to as the \emph{source domain}, while the setting of interest is called the \emph{target domain} \cite{ben2010theory}. Instead of attempting to collect data in an unbiased manner, which might be difficult due to operational, financial or ethical reasons, we are interested in correcting for the domain difference and generalize from the source to the target domain.

In the case of covariate shift, the dominant method of accounting for the differences between domains is \emph{importance-weighting}: samples in the source domain are weighted based on their importance to the target domain. The classifier will subsequently change its predictions in order to avoid misclassifying highly important samples. It has been shown that, under certain conditions, an importance-weighted classifier will converge to the optimal target classifier \cite{cortes2010learning}. How fast it learns depends heavily on how different the domains are, expressed by for instance the R\'enyi divergence \cite{cortes2010learning}. The larger the divergence between the domains, the slower the rate of convergence of the classifier parameter estimator.

Although importance-weighted classifiers are consistent under the right circumstances, their performance still depends strongly on how the importance weights themselves are determined. There has been quite a large variety of work into the behavior of different types of weight estimators: ratio of parametric probability distributions \cite{shimodaira2000improving}, kernel density estimators \cite{silverman1986density}, kernel mean matching \cite{huang2006correcting}, logistic discrimination \cite{bickel2009discriminative}, Kullback-Leibler importance estimation procedure \cite{sugiyama2007covariate}, unconstrained least-squares importance fitting \cite{kanamori2009least}, nearest-neighbour based \cite{loog2012nearest} or conservative minimax estimators \cite{wen2014robust}. Interestingly, these weight estimators can trade off consistency of the estimator for faster convergence, by enforcing smoothness, inhibiting weight distribution bimodality or ensuring a minimum weight value.

Importance-weighting is crucial to evaluating classifiers as well. Model selection is often done through cross-validation, where the training set is split into parts and each part is held back once to be evaluated on later \cite{efron1994introduction,kohavi1995study}. However, the standard cross-validation procedure will not account for domain differences. As a result, its hyperparameter estimates are not optimal with respect to the target domain \cite{sugiyama2005model,sugiyama2007covariate,kouw2016regularization}. Effectively, the standard cross-validation procedure can produce a model selection bias \cite{cawley2010over}. The importance-weighted risk, on the other hand, \emph{can} account for domain differences. By weighting the source validation data, it approximates the target risk more closely. Better approximations of the target risk will allow for hyperparameter estimates that will make the model generalize better.

Importance-weighting is a widely-trusted and influential method, but it can act in quite surprising ways. In this paper we show that, for small sample sizes, the sampling distribution of an importance-weighted estimator can be \emph{skewed}. Skewness refers to the fact that a distribution is not symmetric. That means that, although the estimator is unbiased, it will underestimate the parameter of interest for the majority of data sets, in the case of positive skew. Conversely, it will overestimate the true parameter for the majority of data sets in the case of negative skew. We explore the subsequent effects of this property on model selection under covariate shift.

\section{Preliminaries}
In this section, we introduce our notation, our example setting and explain importance-weighting.

\subsection{Notation}
Consider an input space ${\cal X}$, part of a $D$-dimensional vector space such as $\mathbb{R}^{D}$, and a set of classes ${\cal Y} = \{-1,+1\}$. A source domain is a joint distribution defined over these spaces, $({\cal X}, {\cal Y}, p_{\cal S})$, marked with the subscript ${\cal S}$ and a target domain is another $({\cal X}, {\cal Y}, p_{\cal T})$, marked with ${\cal T}$. Assuming covariate shift implies that the domains' conditional distributions are equal, i.e. $p_{\cal S}(y \mid x) = p_{\cal T}(y \mid x)$, while the marginal data distributions are different, i.e. $p_{\cal S}(x) \neq p_{\cal T}(x)$.

Samples from the source domain are denoted as the pair $(x,y)$, with $n$ samples forming the source dataset ${\cal D}_{\cal S}^{n} = \{(x_i,y_i)\}_{i=1}^{n}$. Similarly, target samples are denoted as $(z,u)$ with $m$ samples forming the target dataset ${\cal D}_{\cal T}^{m} = \{(z_j, u_j)\}_{j=1}^{m}$. A classifier is a function that maps the input space to the set of classes, $h : {\cal X} \rightarrow {\cal Y}$.

\subsection{Example setting} \label{sec:ex}
\begin{figure}[h]
\includegraphics[width=0.48\textwidth]{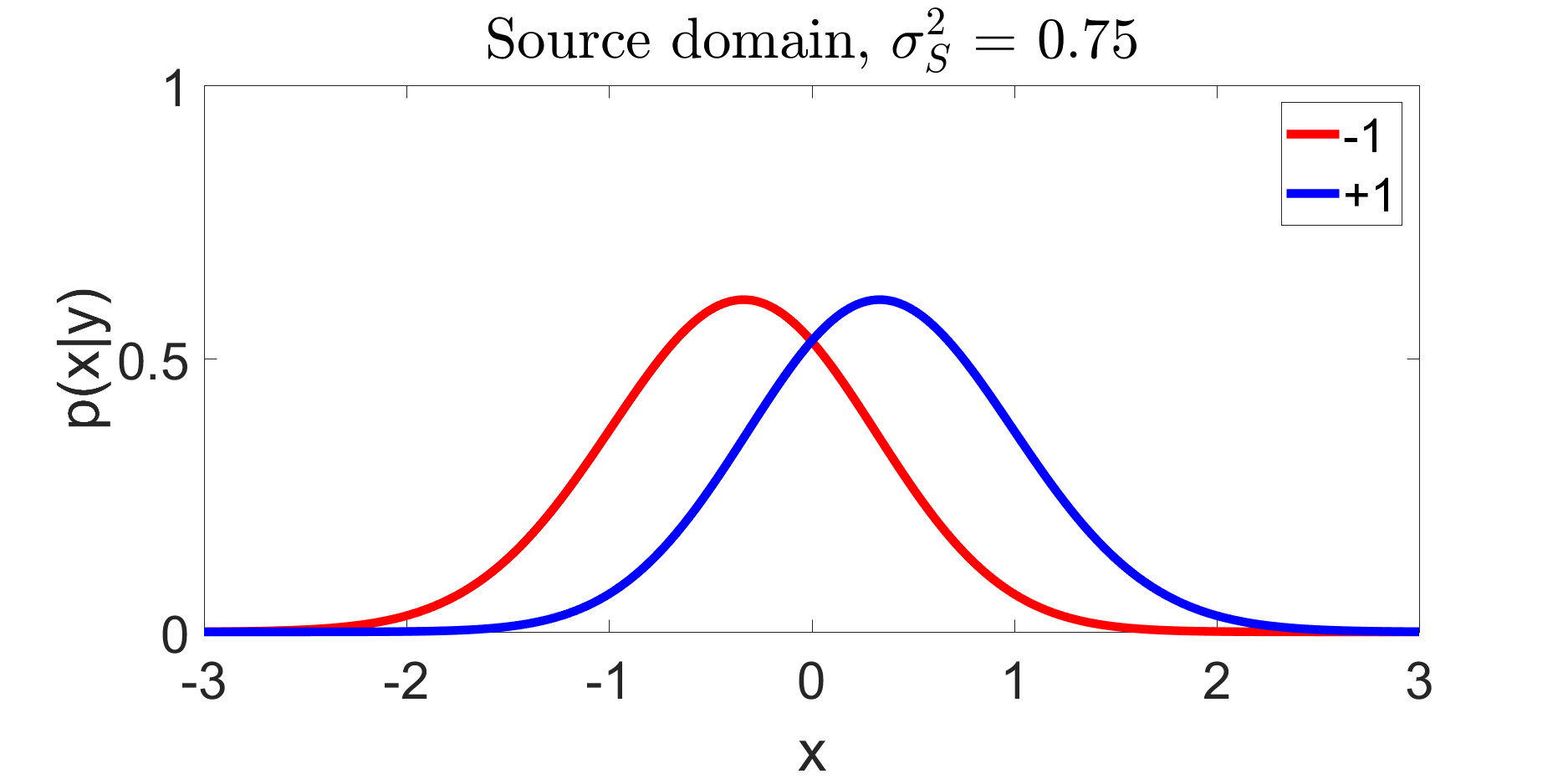} \\
\includegraphics[width=0.48\textwidth]{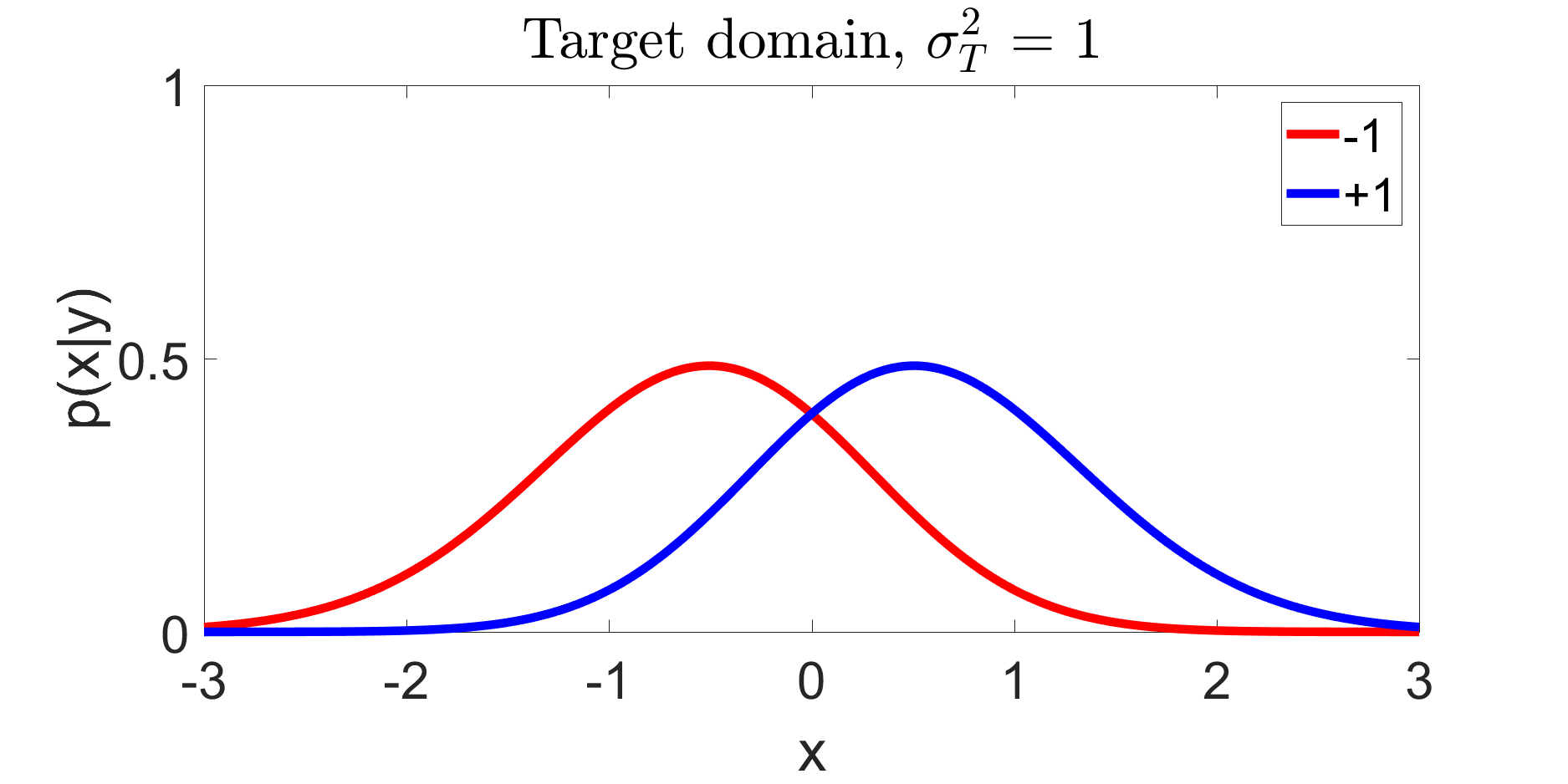}
\caption{Example case of a covariate shift classification problem. (Top) source domain, with $p_{\cal S}(x \mid y) = \Phi(yx)\mathcal{N}(x \mid 0, 0.75) / p_{\cal S}(y)$. (Bottom) target domain, with $p_{\cal T}(x \mid y) = \Phi(yx)\mathcal{N}(x \mid 0, 1) / p_{\cal T}(y)$.}
\label{fig:setting}
\end{figure}

For the purposes of illustrating a few concepts in the upcoming sections, we generate an example of a covariate shift classification problem. For the target data distribution, a normal distribution with a mean of $0$ and a standard deviation of $1$ is taken; $p_{\cal T}(x) = \mathcal{N}(x \mid 0, 1)$. For the source data distribution, we take a normal distribution with a mean of $0$ as well, but with a standard deviation of $0.75$. The class priors in both domains are set to be equal: $p_{\cal S}(y) = p_{\cal T}(y) = 1/2$. Similarly, the class-posterior distributions are set to be equal as well, both in the form of a cumulative normal distribution: $p_{\cal S}(y \mid x) = p_{\cal T}(y \mid x) = \Phi(yx)$. Figure \ref{fig:setting} plots the class-conditional distributions for the source domain (top) and the target domain (bottom). Essentially, the source domain is a biased sample of the target domain, because it favors samples close to $0$ and the decision boundary. Data is drawn through rejection sampling. 

\subsection{Empirical Risk Minimization}
A classifier is a function that assigns a class to each input. Here we will focus on linear classifiers, which project the data onto a vector and make decisons based on which side of the decision boundary the datapoint falls; $h(x) = x\theta$ \cite{friedman2001elements}. In the empirical risk minimization framework, the classifier's decisions are evaluated using a loss function. \emph{Risk} corresponds to the expected loss that the classifier incurs: $R(h)  = \ \mathbb{E}[ \ell(h(x), y) ]$ \cite{mohri2012foundations}. For the examples in this paper, we choose a quadratic loss function, $\ell(h(x), y) =(h(x) - y)^2$ (known for the Fisher classifier and the least-squares SVM). Because the risk function is an expectation, it can be approximated using the sample average $\hat{R}(h)  = \ 1/n \sum_{i=1}^{n} \ell(h(x_i), y_i)$. 

\subsection{Importance weighting}
Considering that each domain has its own joint distribution, it has its own risk function as well. The source risk is $R_{\cal S} = \mathbb{E}_{\cal S} [ \ell(h(x),y)]$, while the target risk is $R_{\cal S} = \mathbb{E}_{\cal S} [ \ell(h(x),y)]$. Their estimators are, respectively:
\begin{align}
	\hat{R}_{\cal S}(h)  =& \ \frac{1}{n} \sum_{i=1}^{n} \ell(h(x_i), y_i ) \nonumber \\
	\hat{R}_{\cal T}(h)  =& \ \frac{1}{m} \sum_{j=1}^{m} \ell(h(z_j), u_j) \nonumber \, .
\end{align}
It is possible to relate the source and target risk functions with each other as follows:
\begin{align}
	R_{\cal T}(h) =& \int_{\cal X} \sum_{y \in {\cal Y}} \ \ell(h(x), y) \ p_{\cal T}(x,y) \  \mathrm{d}x \nonumber \\
	=& \int_{\cal X} \sum_{y \in {\cal Y}} \ \ell(h(x), y) \ \frac{p_{\cal T}(x,y)}{p_{\cal S}(x,y)} p_{\cal S}(x,y) \  \mathrm{d}x \nonumber 
\end{align}
In the case of covariate shift, where $p_{\cal T}(y \mid x) = p_{\cal S}(y \mid x)$, the ratio of the joint distributions $p_{\cal T}(x,y) / p_{\cal S}(x,y)$ can be reduced to the ratio of data marginal distributions $p_{\cal T}(x) / p_{\cal S}(x)$. The new estimator is:
\begin{align}
	\hat{R}_{\cal W}(h) = \frac{1}{n} \sum_{i=1}^{n} \ell(h(x_i), y_i) w(x_i) \nonumber \, ,
\end{align}
where $w(x_i) = p_{\cal T}(x_i) / p_{\cal S}(x_i)$. So, the target risk can be estimated through a weighted average with respect to the source samples. Hence, the ratio of distributions can be recognized as importance weights: the ratio is larger than $1$ for samples that have a high probability under the target distribution relative to the source distribution and smaller than $1$ for samples that have a relatively low probability. 

\begin{figure}[th]
	\includegraphics[width=.45\textwidth]{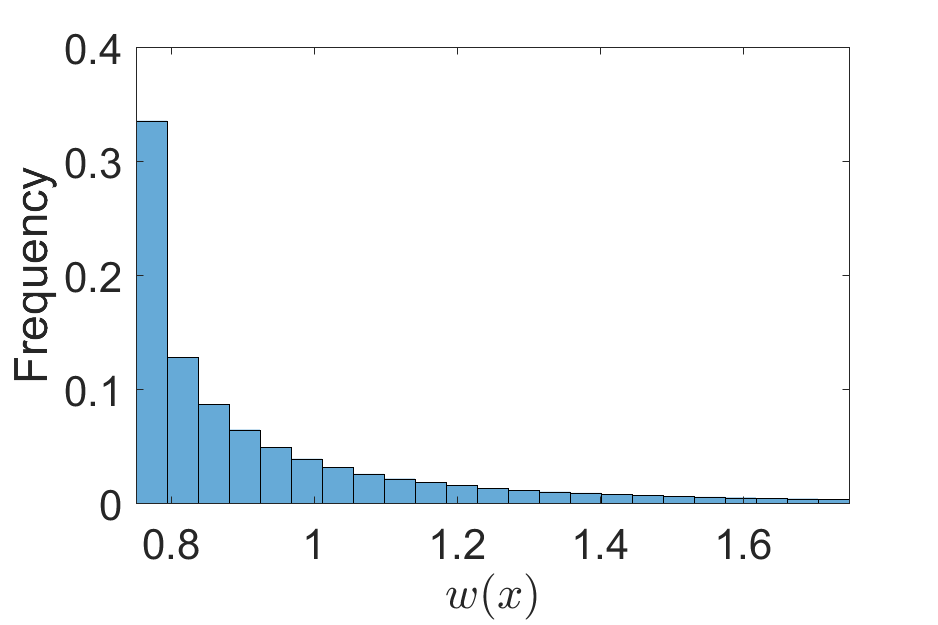}
	\caption{Histogram of the importance weights in the example scenario.}
	\label{fig:histogram_weights}
\end{figure}
The importance weights themselves are often distributed according to an exponential or geometric distribution: many weights are small and a few weights are large. Figure \ref{fig:histogram_weights} presents a histogram for the example setting. As the domains become more dissimilar, eventually, all samples will be nearly zero. 

Weighting can have interesting effects on the behavior of an estimator. The next section discusses the variation in estimates as a function of different data sets. 

\section{Sampling distribution}
The probability distribution of an estimator's results as a function of data, is called the \emph{sampling distribution}. Properties of this distribution are interesting for a number of reasons. Firstly, the difference between the expected value of the sampling distribution and the underlying true risk is called the estimator's bias. It can be desirable to have an unbiased risk estimator: $\mathbb{E}[ \hat{R}(h)] - R(h) = 0$ for all $h$. In other words, there should be no systematic deviation in its estimates. For the case of importance-weighting, it is possible to show that the risk estimator is unbiased:
\begin{align}
	\mathbb{E}_{\cal S}[ \hat{R}_{\cal W}(h)] =& \ \mathbb{E}_{\cal S}[ \frac{1}{n} \sum_{i=1}^{n} \ell(h(x_i), y_i) w(x_i) ]\nonumber \\
		=& \ \frac{1}{n} \sum_{i=1}^{n} \mathbb{E}_{\cal S}[ \ell(h(x), y) w(x) ]\nonumber \\
		=& \ \frac{1}{n} \ n \ \mathbb{E}_{\cal T}[ \ell(h(x), y) ]\nonumber \\
		=& \ R_{\cal T}(h) \nonumber \, . 
\end{align}

\begin{figure*}[th]
\centering
	\includegraphics[width=0.32\textwidth]{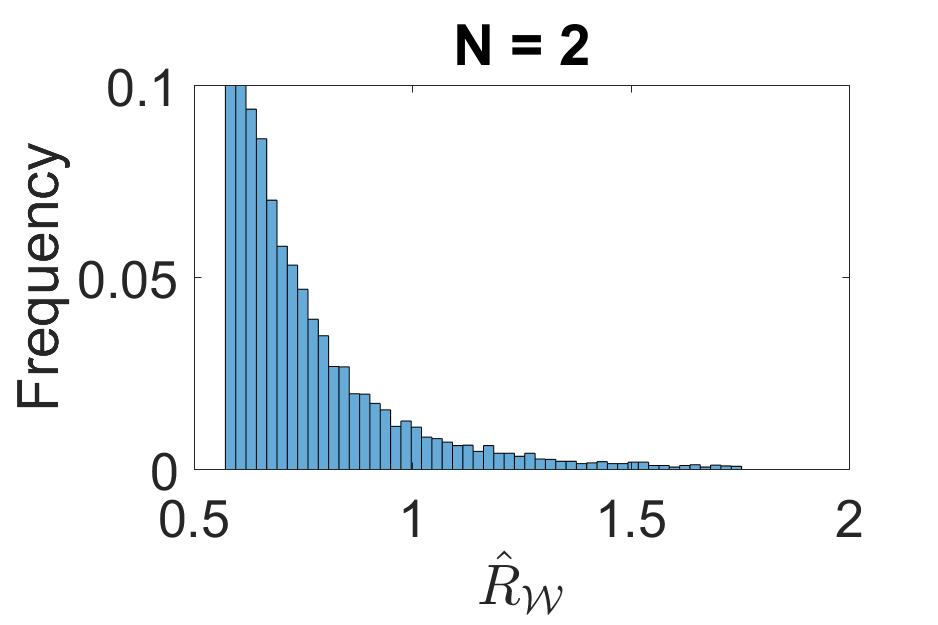}
	\includegraphics[width=0.32\textwidth]{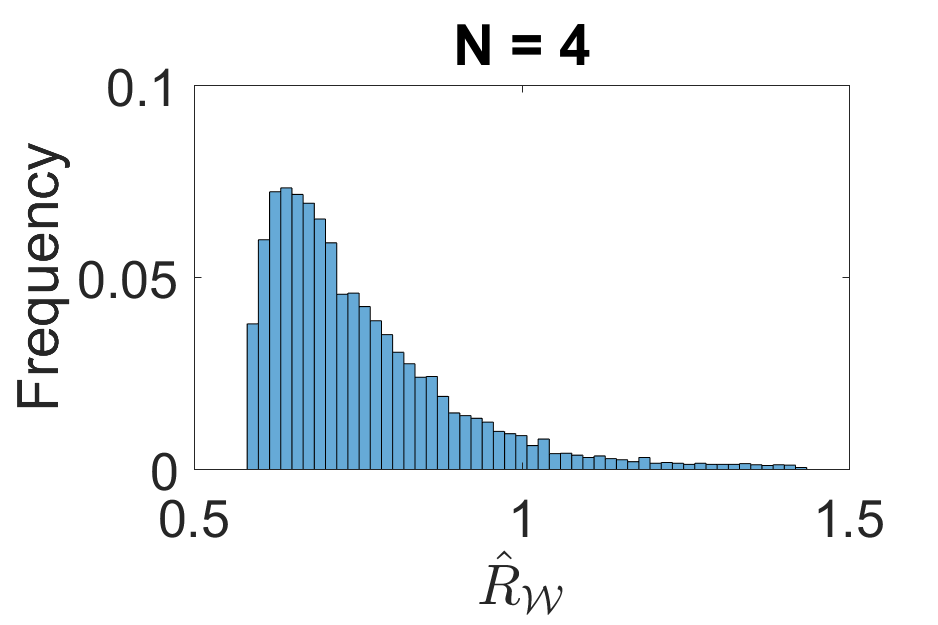}
	\includegraphics[width=0.32\textwidth]{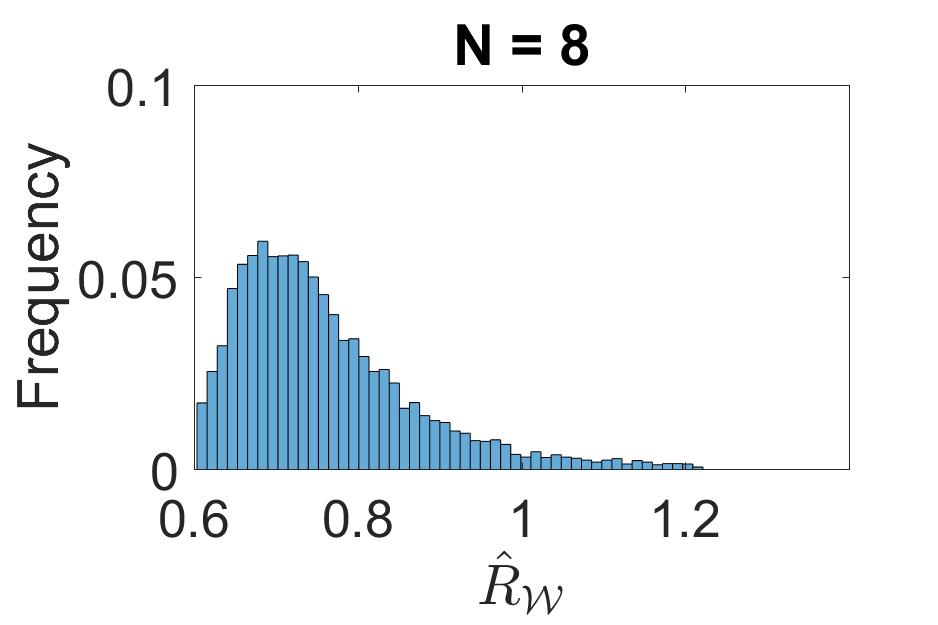} \\
	\includegraphics[width=0.32\textwidth]{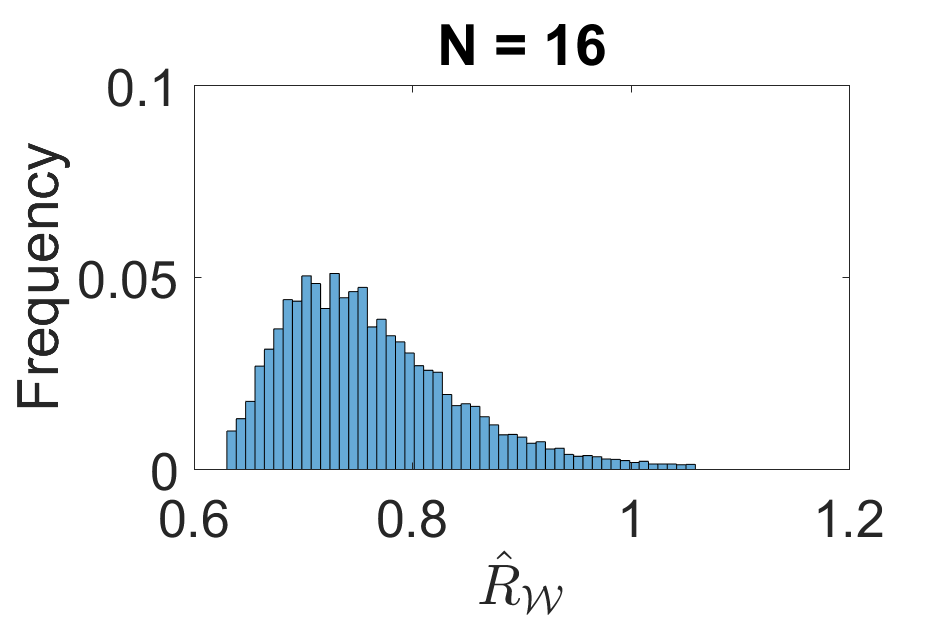}
	\includegraphics[width=0.32\textwidth]{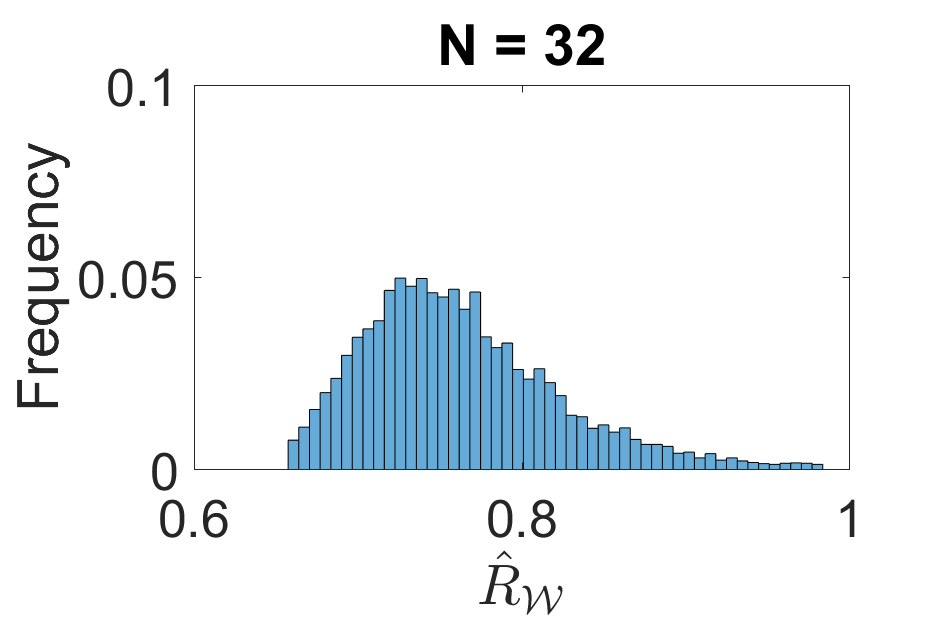}
	\includegraphics[width=0.32\textwidth]{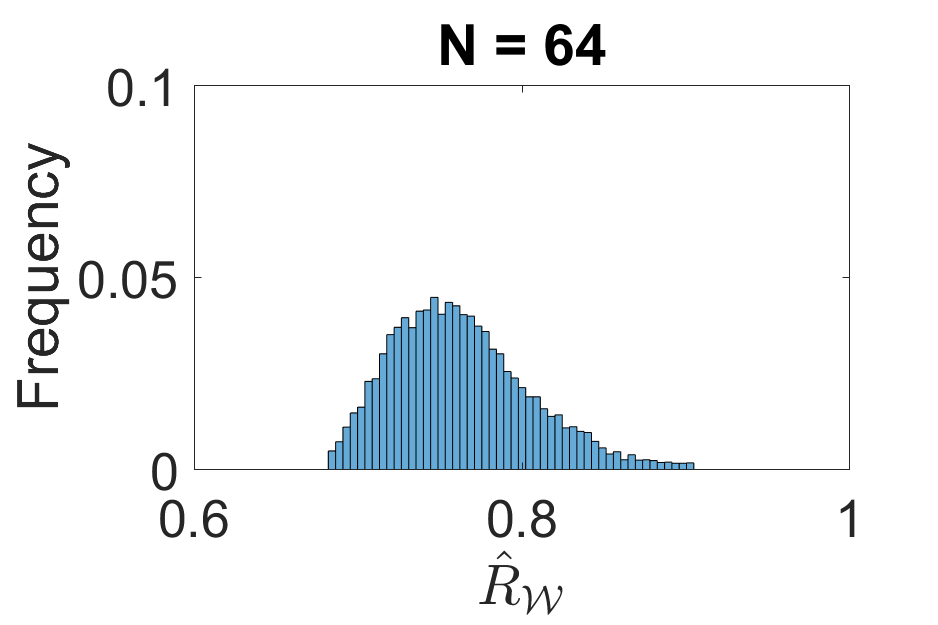}
	\caption{Histograms of the risk estimates $\hat{R}_{\cal W}$ over $10,000$ data sets drawn by rejection sampling from the setting described in Section \ref{sec:ex}, for different sample sizes. Note that the skewness diminishes with more samples.}
	\label{fig:hist_RhW}
\end{figure*}

\subsection{Sampling variance}
Secondly, the variance of the sampling distribution is informative on how uncertain, or conversely how accurate, an estimator is. If the sampling variance reduces as a function of the sample size, then the estimator becomes more accurate with more data \cite{massart2007concentration}. However, in the case of a weighted estimator, depending on the size of the weights, the sampling variance might diverge to infinity \cite{cortes2010learning,kouw2017reducing}. For instance, it can be shown that the variance of the sampling distribution diverges for cases where the domains are too far apart \cite{cortes2010learning}. In fact, for our example case, it can be shown how the weights directly scale the sampling variance:
\begin{align}
	\mathbb{V}_{\cal S}[ \hat{R}_{\cal W}&(h)] = \ \mathbb{E}_{\cal S}[ \Big( \frac{1}{n} \sum_{i=1}^{n} \ell(h(x_i), y_i) w(x_i) - R_{\cal T}(h) \Big)^2]\nonumber \\
=& \ \frac{1}{n^2} \sum_{i=1}^{n} \mathbb{E}_{\cal S}[ \Big( \ell(h(x), y) w(x) - R_{\cal T}(h) \Big)^2]\nonumber \\
		=& \ \frac{1}{n^2} n \ \mathbb{E}_{\cal S}[ \ell(h(x), y)^2 w(x)^2  \nonumber \\
		& \qquad \quad - 2 \ell(h(x), y) w(x) R_{\cal T}(h) \nonumber \\
		& \qquad \quad + R_{\cal T}(h)^2 ] \nonumber \\
		=& \ \frac{1}{n} \Big( \mathbb{E}_{\cal S}[ \ell(h(x), y)^2 w(x)^2]  \nonumber \\
		& \qquad - 2 \mathbb{E}_{\cal S}[ \ell(h(x), y) w(x)] R_{\cal T}(h) \nonumber \\
		& \qquad + R_{\cal T}(h)^2 \Big) \nonumber \\
		=& \ \frac{1}{n} \big( \mathbb{E}_{\cal T}[ \ell(h(x), y)^2 w(x)]  - R_{\cal T}(h)^2 \Big) \nonumber  \, . 
\end{align}
Doing the same derivation for the target risk estimator yields:
\begin{align}
	\mathbb{V}_{\cal T}[ \hat{R}_{\cal T}(h)] = 1/m \big( \mathbb{E}_{\cal T}[ \ell(h(x), y)^2 ]  - R_{\cal T}(h)^2 \Big) \, . \nonumber
\end{align} 
They differ in the expectation term: the weights scale the expected squared loss. For settings where the weights are small, i.e. settings where the domains are close, the importance-weighted estimator converges faster and is more accurate. This fact is exploited in importance sampling \cite{kahn1953methods,neal2001annealed,mcbook}. However, for settings where the weights are large, i.e. settings where the domains are far apart, the weighted estimator has a larger sampling variance, is therefore more uncertain and will need more samples to achieve the same level of accuracy as the target risk estimator. 

\subsection{Sampling skewness}
The \emph{skewness} of a distribution is an indicator of how symmetric it is around its expected value. For small sample sizes, the distribution of the weights can skew the sampling distribution of an importance-weighted estimator. The skewness of a distribution can be expressed using the moment coefficient of skewness: $\Gamma[x] = \mathbb{E}[ \big( (x - \mu)/ \sigma \big)^3 ]$ \cite{cramer2016mathematical,joanes1998comparing}. A negative skew (also known as \emph{left-skewed}) means that the probability mass of the distribution is concentrated to the right of the mean, while a positive skew (a.k.a. \emph{right-skewed}) implies that the probability mass concentrates to the left of the mean. Our importance-weighted estimator is skewed as:
\begin{align}			
	\Gamma_{\cal S} [ \hat{R}_{\cal W}&(h)] = \mathbb{E}_{\cal S} \big[ \Big( \frac{1/n \sum_{i}^{n} \ell((h(x_i), y_i) w(x_i) - R_{\cal T}(h)}{\sqrt{\mathbb{V}_{\cal S} [ \hat{R}_{\cal W}(h)]}} \Big)^3 \big] \nonumber \\
	=& \ \frac{1}{n^3} \sum_{i=1}^{n} \mathbb{V}_{\cal S} [ \hat{R}_{\cal W}(h)]^{-2/3} \nonumber \\
	& \qquad \mathbb{E}_{\cal S} \big[ \big( \ell((h(x), y) w(x) - R_{\cal T}(h) \big)^3 \big] \nonumber \\
	=& \ \frac{n}{n^3} \mathbb{V}_{\cal S} [ \hat{R}_{\cal W}(h)]^{-2/3} \ \mathbb{E}_{\cal S} \big[\ell((h(x), y)^3 w(x)^3 \nonumber \\
	& \qquad -3 \ \ell((h(x), y)^2 w(x)^2 R_{\cal T}(h) \nonumber \\
	& \qquad +3 \ \ell((h(x), y) \ w(x) R_{\cal T}(h)^2 \nonumber \\
	& \qquad - R_{\cal T}(h)^3 \big] \nonumber \\
	=& \ \frac{1}{n^2} \mathbb{V}_{\cal S} [ \hat{R}_{\cal W}(h)]^{-2/3} \ \Big( \mathbb{E}_{\cal T}[\ell((h(x), y)^3 w(x)^2] \nonumber \\
	& \qquad -3 \ R_{\cal T}(h) \mathbb{V}_{\cal S}[\hat{R}_{\cal W}(h)] \nonumber \\
	& \qquad - R_{\cal T}(h)^3 \Big) \label{eq:skew} \nonumber \, .
\end{align}

Again, doing the same derivation for the target risk estimator leads to:
\begin{align}
	\Gamma_{\cal T}[ \hat{R}_{\cal T}(h)] =& \ \frac{1}{m^{2}} \ \mathbb{V}_{\cal T} [ \hat{R}_{\cal T}(h)]^{-2/3} \ \big( \mathbb{E}_{\cal T}[\ell((h(x), y)^3] \nonumber \\
	& \qquad -3 \ R_{\cal T}(h) \mathbb{V}_{\cal T}[\hat{R}_{\cal T}(h)] - R_{\cal T}(h)^3 \big) \, , \nonumber 
\end{align}
showing that the skew of the importance-weighted estimator depends on multiplying the cubic loss with the squared weights. If the weights are large, the existing skew is scaled up. Note that the skew also reduces as the sampling variance increases.

The moments of the sampling distribution of the risk estimator depend heavily on the problem setting. It is therefore difficult to make general statements regarding all possible covariate shift classification problems. We can, however, illustrate the skew for the example case. In order to evaluate the risk estimator's ability to validate a classifier, the classifier needs to remain fixed while the risk function is computed for different data sets. We took a linear classifier, $h(x \mid \theta) = x\theta$ with $\theta = 1/(2\sqrt{\pi})$. Figure \ref{fig:hist_RhW} plots the histograms of $10,000$ repetitions of rejection sampling. Note that each repetition correponds to a single validation data set. After computing the risks, it becomes apparent that the sampling distribution of $\hat{R}_{\cal W}$ is positively skewed and that its skew diminishes as the sample size increases.

\begin{figure*}[th]
\centering
	\includegraphics[width=0.32\textwidth]{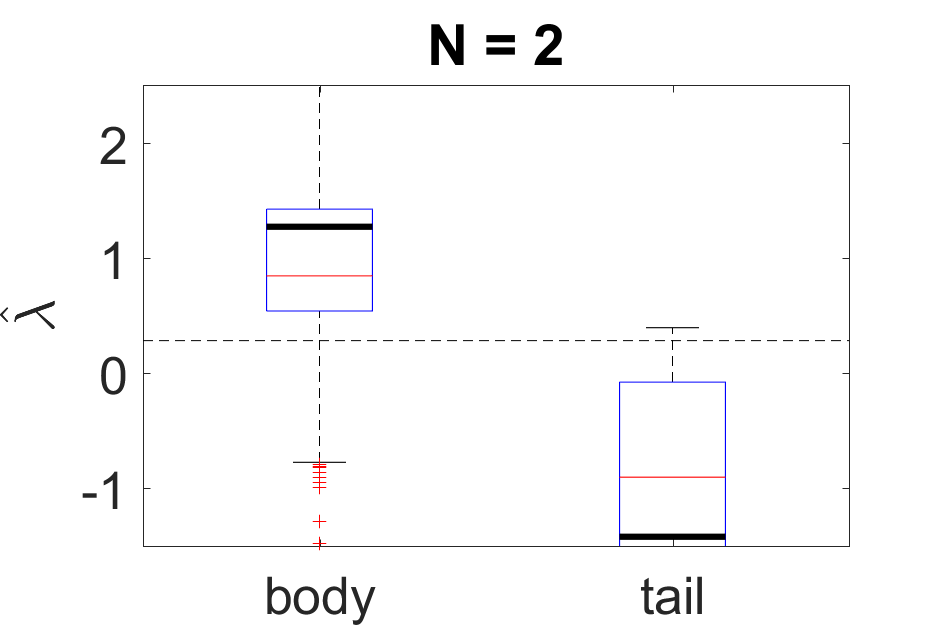}
	\includegraphics[width=0.32\textwidth]{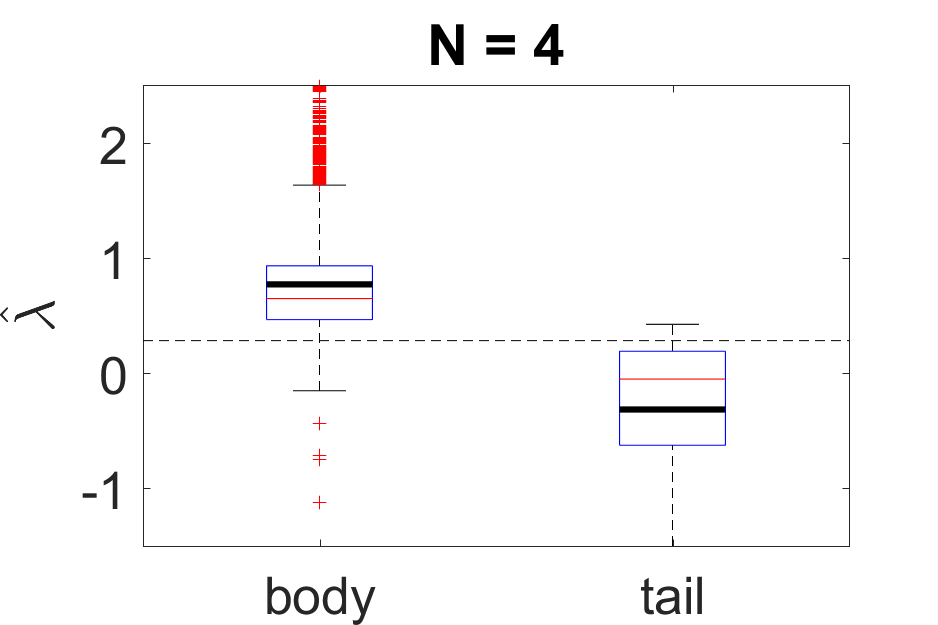}
	\includegraphics[width=0.32\textwidth]{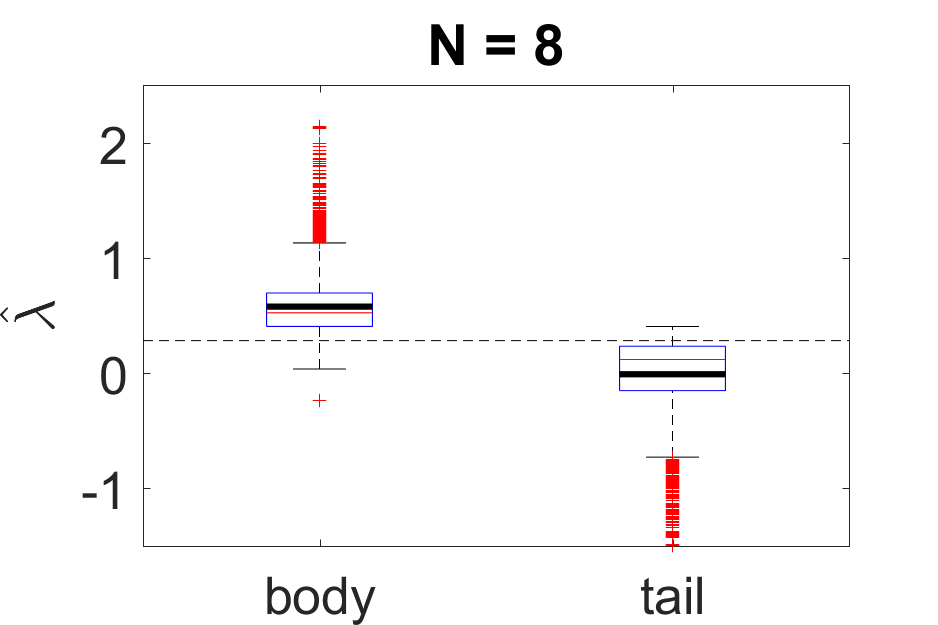}
	\includegraphics[width=0.32\textwidth]{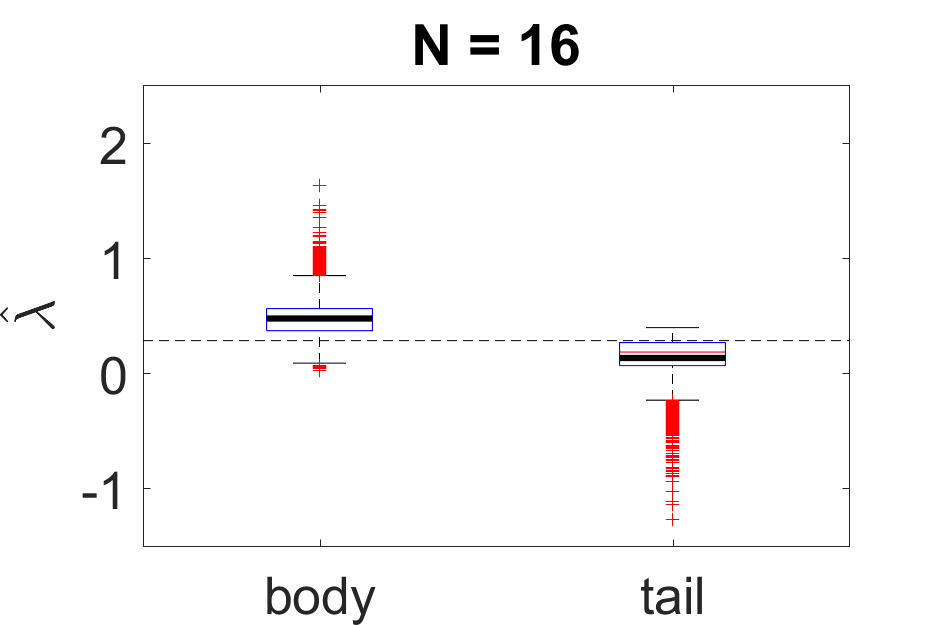}
	\includegraphics[width=0.32\textwidth]{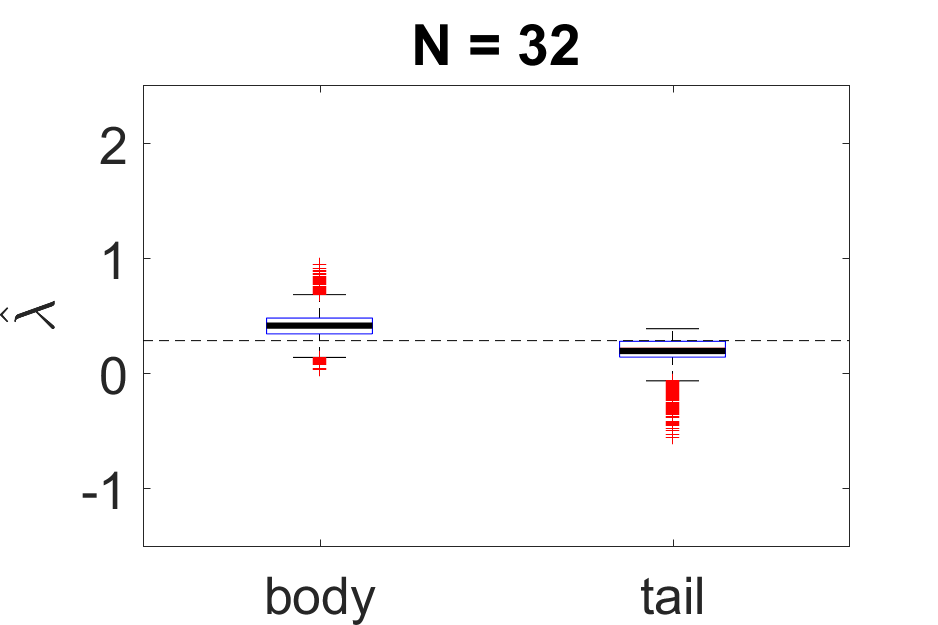}
	\includegraphics[width=0.32\textwidth]{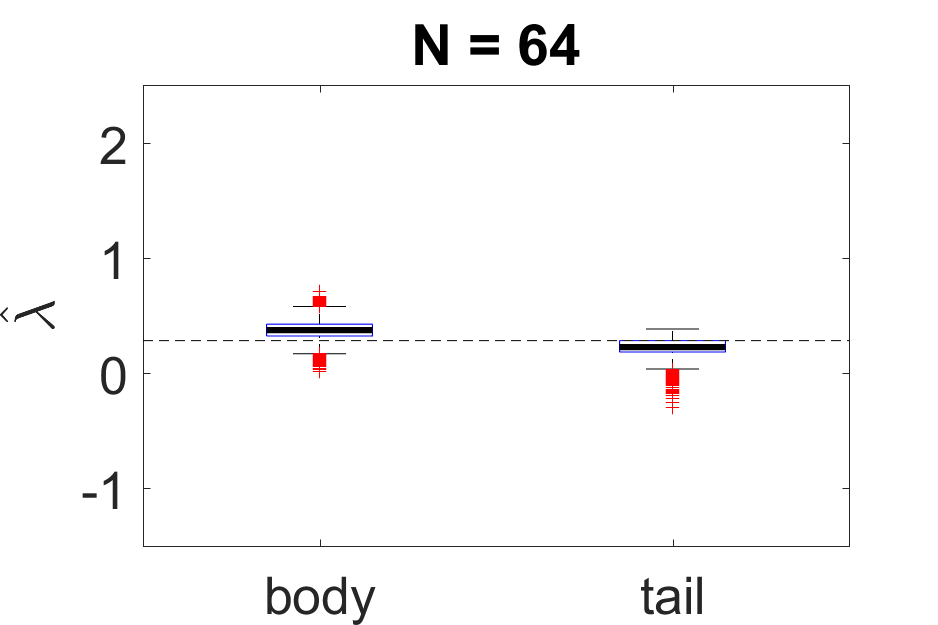}
	\caption{Boxplots of the regularization parameter estimates $\hat{\lambda}$ based on the importance-weighted risk estimator $\hat{R}_{\cal W}$, for different sample sizes.}
	\label{fig:box_bodyvstail}
\end{figure*}

\section{Model selection}
The importance-weighted risk estimator is crucial to model selection under covariate shift. Standard cross-validation does not account for domain differences \cite{sugiyama2005model}. Validating the model on source samples leads to hyperparameters that are not optimal with respect to the target domain \cite{kouw2016regularization}. Importance-weighting the source data results in a validation data set that more closely matches the target domain \cite{sugiyama2007covariate}. However, importance-weighed cross-validation suffers from a number of issues: for large domain differences, the sampling variance can diverge, resulting in highly inaccurate estimates \cite{cortes2010learning,kouw2017reducing} and for small sample sizes, the sampling distribution can be skewed. How this skew affects validation will be shown in the following experiment. 

\subsection{Body versus tail}
The defining property of a skewed distribution is that the majority of the probability mass lies to the side of its expected value. The narrow region with a large amount of probability mass is called the \emph{body}, while the long low-probability-mass region to the side of the body is called the \emph{tail}. In the case of the example setting, the weighted risk estimator's sampling distribution has a body on the left and a tail that drops off slowly to the right, as can be seen in Figure \ref{fig:hist_RhW} for $N=2$. Note that high probability mass regions of a sampling distribution correspond to many data sets. The risk estimates in the body are smaller than that of the expected value of the sampling distribution, i.e., the true target risk $R_{\cal T}$. Hence, the body contains \emph{under}estimates of the target risk. The right-hand tail on the other hand contains \emph{over}estimates. Note that the body contains many, relatively small, underestimates while the tail contains a few, relatively large, overestimates. We know that they cancel out, because we know the importance-weighted risk estimator is unbiased. However, \emph{for the large majority of data sets}, the risk is underestimated. This directly affects the hyperparameter estimates obtained in cross-validation.

\subsection{Regularization parameter selection}
In order to evaluate the importance-weighted risk estimator's usefulness for model selection, we evaluate it for a regularized classifier. The problem setting is still the the example setting from Section \ref{sec:ex}. We take the same linear classifier as before, but this time we add $L^2$-regularization: $h(x) = x \theta_{\lambda}$ where $\theta_{\lambda} = 1/(2\sqrt{\pi})+\lambda$. We draw $N$ samples from the source domain and evaluate $\theta_{\lambda}$ using the importance-weighted risk estimator. Following that, we select the $\lambda$ for which the risk is minimal: $\hat{\lambda} = \ \arg \min_{\lambda} \  \hat{R}_{\cal W}(\theta_{\lambda})$. For the example case, $p_{\cal T}(x,y) = \Phi(yx) \mathcal{N}(x \mid 0,1)$, the expected risk can be found analytically. The true risk is minimal for $\theta_{\lambda} = 1 / \sqrt{\pi}$, which means the optimal value of $\lambda$ is $1/(2\sqrt{\pi})$. The better the risk estimator approximates the expected risk, the better the resulting $\hat{\lambda}$ will approximate the optimal value for $\lambda$. 

The above procedure of drawing samples, computing risk and selecting $\lambda$ is repeated $10,000$ times. All data sets for which the risk is smaller than the average risk over all repetitions are deemed part of the body, while all data sets with risks larger than the average are deemed part of the tail. Figure \ref{fig:box_bodyvstail} shows the boxplots of $\hat{\lambda}$ for the body and tail separately. Each subplot covers one sample size: $N = \{2,4,8,16,32,64\}$. The dotted line corresponds to the optimal $\lambda$, and the black bars in the boxplots mark the average estimates. For $N=2$, the body produces overestimates of the regularization parameter on the order of $+1$, while the tail produces underestimates on the order of $-2$. For $N = 4$ to $8$, the effect is smaller, with the tail producing more accurate estimates. For $N \ge 32$, the differences between the body and the tail are nearly gone.

\begin{figure}[h!]
\centering
	\includegraphics[width=0.48\textwidth]{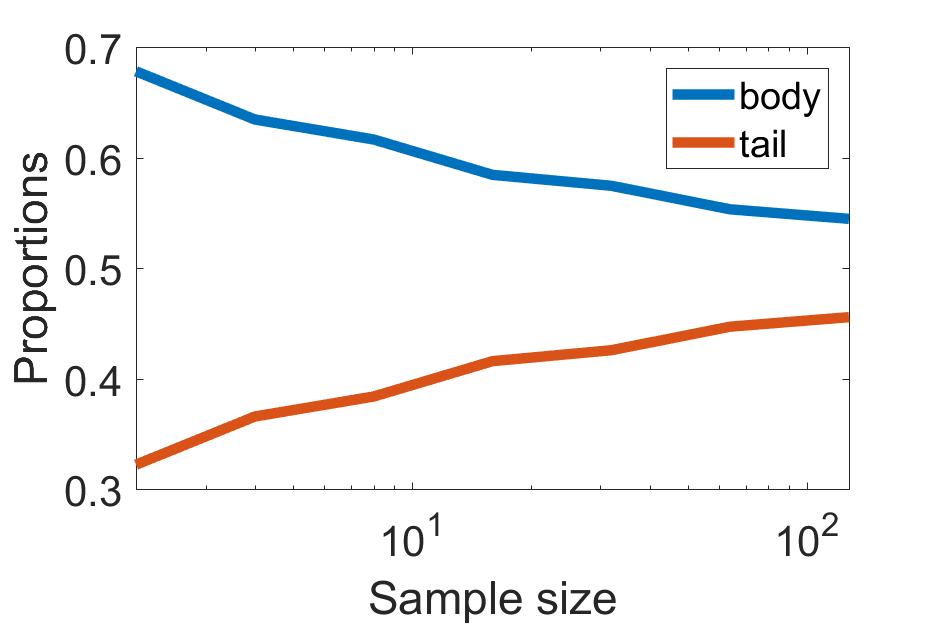}
	\caption{Proportions of data sets in the body (blue) versus tail (red).}
	\label{fig:prps}
\end{figure}

Figure \ref{fig:prps} plots the proportions of data sets belonging to the body versus the tail, out of the $10,000$ repetitions. Looking at the amount of data sets that make up each part, we can conclude that the majority is part of the body. For the smallest data sets, there are, in fact, twice as many data sets in the body. As the sample size increases, the sampling distribution becomes less skewed and the proportions become equal. 


\section{Discussion}
Although the current problem setting is $1$-dimensional, we do believe that higher-dimensional problem settings behave along the same lines. However, the current indications of ''enough'' validation data may not be completely indicative of higher-dimensional settings; we expect that more validation data is required in that case. Also, in the reverse setting, where the source domain is wider than the target domain, the skew of the risk estimator's sampling distribution is negative instead of positive, which means that all statements regarding over- and underestimates are reversed. 

We have chosen a quadratic loss function to evaluate risk, but we believe that the results presented here will hold for other choices of loss functions as well. The skewness stems from the skewness of the importance-weights, which does not depend on the loss function. 

A limitation of our study is the fact that it only covers the case of Gaussian data distributions. It would be helpful to attain a more general understanding of the effects of the skewness of the risk estimator's sampling distribution. Unfortunately, generalizing the behavior of a sampling distribution for all possible covariate shift problem settings is not trivial. 

Nonetheless, if the skew in the sampling distribution is indeed caused by the geometric distribution of the weights, then the current results might extend to importance-weighted classifiers as well. If the sampling distributions of classifier parameter estimators are skewed, then we might also see many overestimates of classifier parameters for the majority of data sets and a few large underestimates for rare cases, when sample sizes are small. Regularization has the potential to correct this, which again stresses the importance of having a good model selection procedure for covariate shift problems.

\section{Conclusion}
We presented an empirical study of the effects of a skewed sampling distribution for the importance-weighted risk estimator on model selection. Depending on the problem setting, for small sample sizes, the estimator will frequently produce underestimates and infrequently produce large overestimates of the target risk. When used for validation, the risk estimator ensures that the regularization parameter is overestimated for the majority of data sets, for cases of sample selection bias. However, with enough data, the skew diminishes.





\bibliographystyle{IEEEtran}
\bibliography{kouw_icpr18a}

\end{document}